\documentclass{elsarticle}
\usepackage{setspace}
\usepackage{caption}
\usepackage{lineno,hyperref}
\usepackage{graphicx}
\graphicspath{{./images2/}}
\usepackage{comment}
\usepackage{amsmath,amssymb} 

\DeclareMathOperator*{\argmax}{argmax}
\DeclareMathOperator*{\argmin}{argmin}
\newcommand{\RNum}[1]{\uppercase\expandafter{\romannumeral #1\relax}}
\usepackage{amsfonts}
\usepackage{mathdots}
\usepackage{bbm}
\usepackage{amssymb}
\usepackage{makeidx}
\usepackage{mathtools}
\usepackage{amsthm}
\usepackage{algorithm}
\usepackage{algpseudocode}
\usepackage{multirow}
\usepackage{multicol}
\usepackage{tabu}
\usepackage{arydshln}
\usepackage{tabularx}
\usepackage{booktabs}
\usepackage{makecell}
\modulolinenumbers[5]
\usepackage{pifont}
\usepackage{xcolor}

\journal{Pattern Recognition \LaTeX\ Templates}

\begin{document}
\begin{frontmatter}

\title{Orthonormal Product Quantization Network for Scalable Face Image Retrieval}

\author[1,2,3]{Ming Zhang\corref{mycorrespondingauthor}}
\cortext[mycorrespondingauthor]{Corresponding author}
\ead{mzhang367-c@my.cityu.edu.hk}

\author[4]{Xuefei Zhe}
\ead{elizhe@tencent.com}

\author[1,2]{Hong Yan}
\ead{h.yan@cityu.edu.hk}

\address[1]{Department of Electrical Engineering, City University of Hong Kong, Hong Kong}
\address[2]{Centre for Intelligent Multidimensional Data Analysis Limited, Hong Kong}
\address[3]{Hong Kong Applied Science and Technology Research Institute Company Limited, Hong Kong}
\address[4]{Tencent AI Lab, Shenzhen, China}

\begin{abstract}
Existing deep quantization methods provided an efficient solution for large-scale image retrieval. However, the significant intra-class variations, like pose, illumination, and expressions in face images, still pose a challenge. In light of this, face image retrieval requires sufficiently powerful learning metrics, which are absent in current deep quantization works. Moreover, to tackle the growing unseen identities in the query stage, face image retrieval drives more demands regarding model generalization and scalability than general image retrieval tasks. This paper integrates product quantization with orthonormal constraints into an end-to-end deep learning framework to effectively retrieve face images. Specifically, we propose a novel scheme that uses predefined orthonormal vectors as codewords to enhance the quantization informativeness and reduce codewords' redundancy. A tailored loss function maximizes discriminability among identities in each quantization subspace for both the quantized and original features. An entropy-based regularization term is imposed to reduce the quantization error. Experiments are conducted on four commonly-used face datasets under both seen and unseen identity retrieval settings. Our method outperforms all the compared state-of-the-art under both settings. The proposed orthonormal codewords consistently boost both models' standard retrieval performance and generalization ability, demonstrating the superiority of our method for scalable face image retrieval.
\end{abstract}


\end{frontmatter}
\nolinenumbers
\section{Introduction}
Rapid growth in the internet user population and the popularity of mobile devices with advanced cameras have prompted the sharing of visual content on social media. A large number of user-generated human face images, e.g., selfies and portraits, are uploaded every day. Due to the need for image indexing and searching, large-scale image retrieval~\cite{yu2014learning,zhao2022feature} has been an active area of research. Face image retrieval~\cite{tang2018discriminative, zaeemzadeh2021face, zhang2021deep} aims to return images from the database images that are of the same person as the query image. However, large intra-class variances caused by expressions, illumination, or occlusion, and small inter-class distances between visually similar people, make developing an accurate and efficient system for unconstrained face image retrieval challenging. Another problem is, in a real-world application where the number of newly joined identities keeps growing, high scalability of the retrieval system is needed. The poor generalization performance prevents the scaling of these retrieval systems to larger datasets.

One basic idea for highly efficient image retrieval is using binary code representations to encode the data, thereby enabling an approximate nearest neighbor (ANN) search to accelerate the query process. According to their applied retrieval metrics, the works of obtaining binary code representations can be divided into two types: 1) Hamming distance- or 2) dictionary-related distance-based. Following practice in the literature~\cite{klein2019end,jang2020generalized,yu2020product}, we refer to Hamming distance-based approaches as \textbf{\textit{hashing}} models and dictionary-related distance-based approaches as \textbf{\textit{quantization}} models. The goal of hashing is to map high dimensional real-valued data to lower dimensional binary codes in Hamming space while preserving their original similarity. The Hamming distance of the binary codes between the query and database images can be computed extremely fast by the XOR operation. 
Recently, supervised deep hashing~\cite{li2016feature,lin2017discriminative,Yuan_2020_CVPR}, using deep convolutional neural networks (CNNs), was proposed for end-to-end learning of feature representations and hashing functions and substantially outperformed traditional hashing methods~\cite{song2018quantization} for image retrieval. 

For any pair of binary codes of length $l$, hashing-based methods can only generate $l+1$ distinct values to depict their pairwise similarity making it hard to draw rich similarity relations for large-scale face image datasets with many classes of identities. Another disadvantage is most hashing methods obtain binary codes by applying a sign function to continuous features~\cite{li2017deep,lin2017discriminative}. To solve the intractable discrete optimization in the training process, they usually relax the discrete constraint to be continuous and convert it to a regularization term. Consequently, it causes inevitable information loss. 

In parallel with binary hashing, product quantization (PQ)~\cite{jegou2010product, ge2013optimized, zhang2014composite} has been widely employed in the fields of computer vision and information retrieval. It decomposes feature vectors in the original space into several disjoint sub-vectors. Each sub-vector then needs to find the nearest centroid (codeword) in the subspace (codebook). By replacing each sub-vector with the index of the nearest codeword, the original features in one subspace are encoded into binary codes. For binary codes of length $l=M \log_2 K$, where $M$ is the number of codebooks, and $K$ is the number of codewords in each codebook, PQ is capable of producing $\tbinom{K}{2}^M$ distinct distance values. Thus, it is more powerful to describe the similarity distance between fine-grained face samples. During the query stage, PQ-based methods allow the use of multiple look-up tables (LUTs) for query speed acceleration, which is only slightly more costly than hashing-based methods~\cite{jegou2010product, cao2016deep}. Although deep hashing has drawn increasing attention for face image retrieval~\cite{tang2018supervised,tang2018discriminative,zhang2021deep}, deep quantization methods are rarely publicly explored for the task.

The PQ technique was initially designed under an unsupervised setting. 
Recently, some deep quantization methods~\cite{yu2018product,klein2019end} have been proposed to learn codewords with supervision. The feature representations are divided into several sub-vectors and quantized by learnable codewords depending on the similarity between the sub-vectors and codewords. Typically, a softmax or triplet loss can be constructed based on the resulting quantizations. Thereby, the learning metrics preserve the label information in both the feature representations and the learnable codewords. We refer to these deep quantization methods with the learning manner as \textit{\textbf{learning to quantization (l2q)}}. Nevertheless, since the codewords themselves do not contain any discriminative information for the quantization process, it is possible to decorrelate the discriminative visual information and codewords individually. Under this hypothesis, we can learn the codewords assignment of feature representations even with predefined codewords. 

\begin{figure}[htbp]
	\centering
	\includegraphics[width = 0.6\textwidth]{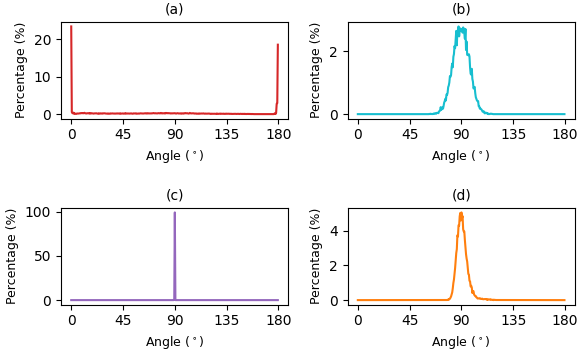}
	\caption{Distribution of angles between pairs of codewords in different deep quantization models: (a) DPQ~\cite{klein2019end}, (b) GPQ~\cite{jang2020generalized}, (c) the proposed OPQN method, and (d) the variant of OPQN without predefined codewords. Each angular distribution is a normalized histogram, generated by quantization from $0^\circ$ to $180^\circ$, step by $0.5^\circ$.}
	\label{ang_dst}
\end{figure}

From the view of codewords, how codewords' distribution has an influence on the quantization quality is seldom explored previously. Firstly, codewords in a codebook can be regarded as prototypes in a subspace that spread over the subspace but should retain some distance from each other. 
To visualize how codewords are distributed in l2q methods, we compute the angles between codeword pairs in each codebook in two deep quantization models~\cite{klein2019end,jang2020generalized} and illustrate their distributions in Fig.~\ref{ang_dst}(a) and (b). We can see that the compared methods show variable angular distributions, which are sub-optimal in terms of distinct codewords separation and low codewords redundancy. Secondly, hand-crafted codewords impose explicit constraints directly on the codewords distribution. Considering the orthogonal case, the designed codewords will exhibit a fixed decent 90-degree separation from each other. Last but not least, predefined codewords enable reusing identical codebooks for different datasets, helping to lower the system's storage cost and reduce preprocessing time.

Motivated by the aforementioned codeword analysis, we argue that codewords' distribution plays an important role in quantization learning. Specifically, codewords that scatter uniformly in the subspace and keep a distinct distance from each other benefit both the model's quantization performance and generalization ability. It is noteworthy that validation on model generalization to unseen data is largely neglected by prior deep hashing-based face image retrieval works~\cite{lin2017discriminative,zhang2021deep}. Consequently, prior works only limit their evaluation protocol to seen identities so that the queries used for evaluation share the same identity classes as the training set. However, hashing models performing well under this setting may work badly for unseen identity retrieval, as reported in~~\cite{sablayrolles2017should}. To this end, this paper proposes a method called Orthonormal Product Quantization Network (OPQN) to explicitly design codewords and employ them to learn alternative parameters for quantization. OPQN predetermines sets of orthonormal vectors as codewords instead of learning them for use. Thus, it exhibits the angular distribution of 90-degree between each pair of codewords, as illustrated in Fig.~\ref{ang_dst}(c). Since the procedure is the opposite of the conventional l2q methods, OPQN belongs to \textbf{\textit{quantization to learning}}. 
Equipped with these predefined codewords, OPQN learns more separable features in the hyper-sphere space. To further enhance the competence of the learning metric, an angular margin-based loss is proposed which makes the most of the discriminability of feature representations in each subspace. The main contributions of this paper are summarized as followings:
\begin{itemize}
\item We propose a novel deep quantization method producing compact binary codes for large-scale face image retrieval. Our method uses predefined orthonormal vectors as codewords to increase quantization informativeness and reduce codeword redundancy. Besides, it has lower storage costs and more efficient asymmetric comparisons in the retrieval phase.

\item We design a tailored loss function, which maximizes the discriminability of identities in each subspace. It works simultaneously on the original and the quantized features to give a better quantization quality. We also impose an entropy-based regularization term to boost the retrieval performance under tiny bits.

\item 
Extensive experiments show that OPQN outperforms all compared state-of-the-art and generalizes the best to retrieve unseen identities. Combining with further experiments for general image retrieval\footnote{Experiment results on two general image datasets are shown in the supplementary material.}, it demonstrates the broad superiority of the proposed codewords scheme and the learning metric, providing a general framework for deep quantization-based scalable image retrieval.
\end{itemize}

Following in this paper, Section~\ref{sec:2}, recalls prior works related to our approach. The proposed OPQN method, including the generation of codewords and design of the loss function, is described in Section~\ref{sec:3}. In Section~\ref{sec:4}, OPQN is evaluated on four benchmark datasets: FaceScrub~\cite{ng2014data}, CFW-60K~\cite{li2015two}, VGGFace2~\cite{cao2018vggface2}, and YouTube Faces~\cite{wolf2011face} under both seen and unseen identity retrieval settings. In Section~\ref{sec:5}, an ablation study with elaboration is reported and the performance of the model with respect to codebook configurations and parameter sensitivity is discussed. 
The paper's conclusions and outlook for future research are presented in Section~\ref{sec:6}.

\section{Related work} \label{sec:2}
In this section, some representative works on supervised deep hashing are reviewed first. Then, traditional quantization methods and state-of-the-art deep quantization methods for image retrieval are introduced.

\subsection{Supervised Deep Hashing for Image Retrieval}
Based on the approach of labels utilization, existing supervised deep hashing methods can be roughly classified into three types: pairwise label-based~\cite{li2016feature,fu2022deep}, triplet label-based~\cite{wang2016deep,yao2016deep}, and class-wise label-based~\cite{zhe2019deep,Yuan_2020_CVPR}. It is known that pairwise label-based methods cannot capture the complete similarity information underlying the dataset, and the triplet label-based methods suffer from high computational costs. To address these, 
recently, some methods~\cite{zhe2019deep,Yuan_2020_CVPR,zhang2021improved} using class-wise label-based similarity were developed, which can generate more discriminative and compact hashing codes. More recently, DCGH~\cite{zhang2023deep} proposed a collaborative graph hashing framework, which utilized multi-level semantic information across visual and semantic spaces. DCGH built a graph neural network to retain the latent structural relations in the learned hashing codes.

Previous deep hashing works on face image retrieval mainly focused on the design of the network architecture and widely adopted softmax classification loss for supervision. Specifically, a fully connected (FC) layer transformed bottleneck features to hashing outputs in Euclidean space, which were usually supervised by a softmax classifier. Generally, a quantization loss was also imposed to relax discrete binary constraints to be continuous and to reduce the quantization error. In Discriminative Deep Hashing (DDH)~\cite{lin2017discriminative}, deep CNNs were trained with a divide-and-encode module to obtain compact binary codes for face image retrieval. Following DDH, Discriminative Deep Quantization Hashing (DDQH)~\cite{tang2018discriminative} found that retrieval performance can be further enhanced by inserting a batch normalization layer between the FC layer and the Tanh activation function. Recently, several works utilized label information with other supervisions. 
Inspired by the class-wise label-based similarity~\cite{zhe2019deep}, \cite{zhang2021deep} proposed Deep Center-based Dual-constrained Hashing (DCDH). It used a center-based framework to jointly learn hashing functions and class centers end-to-end, achieving state-of-the-art results on face image retrieval. However, the method cannot necessarily capture the underlying semantic similarity of datasets. This limitation is reflected in the model's poor generalization performance for unseen identity retrieval in Section~\ref{cross-domain}. 

\subsection{Traditional Quantization Techniques}
Vector Quantization (VQ) is the most classical quantization technique, which quantizes the feature space by maintaining one codebook. Suppose that the codebook consists of $K$ codewords, VQ divides the feature space into $K$ clusters using unsupervised clustering methods so that each feature vector can be encoded by $log_2 K$ bits. The LUT stores the pre-computed distance matrix between every two clusters and has $\mathcal{O}(K^2)$ entries. An increase in VQ's bit length will lead to exponential growth in the number of clusters, $K$, and the number of entries in LUT grows quadratically with $K$, making the method impractical for large values of $K$ and restricting its utility. PQ~\cite{jegou2010product} overcomes this limitation. It decomposes a feature vector $x_{i}\in \mathbb{R}^{Md}$ into $M$ disjoint sub-vectors with dimension $d$, i.e. $x_i = [x_{i1}, x_{i2},\cdots,x_{iM}]$. The sub-vector $x_{im}$ is related to the $m$-th subspace, which is quantized by the codebook $C_m = [C_{m1}, C_{m2},\cdots, C_{mK}] \in \mathbb{R}^{d \times K}$, composed of $K$ codewords. By representing a feature vector with $M$ codebooks, PQ can achieve $K^M$ combinations of codeword. Therefore, it outperforms VQ with more expressive power for quantization. Optimized PQ methods, such as AQ~\cite{babenko2014additive} and CQ~\cite{zhang2014composite} have been developed to achieve a more accurate decomposition of the feature space and learning of codewords. 

\subsection{Deep Quantization for Image Retrieval}
Recently, deep quantization methods have emerged as an effective solution for image retrieval tasks, integrating quantization into deep CNNs for simultaneous feature learning and codeword learning. Deep Quantization Network (DQN)~\cite{cao2016deep} was the first attempt of this kind, which introduced a combined similarity-preserving and product quantization loss.
Deep Product Quantization (DPQ)~\cite{klein2019end} learned both soft and hard quantizations for a more accurate asymmetric search. It applied a straight-through (ST) estimator to enable back-propagation (BP) on the $\argmax(\cdot)$ function. More recently, \cite{yu2018product} proposed a product quantization network (PQN), which used a soft PQ layer to directly determine codeword assignments from the cosine similarity between features and codewords. Specifically, both $x_{im}$ and $C_{mk}$ are $\ell_2$ normalized to unit length so that their similarity can be taken directly from their inner product. The soft quantization $s_{im}$ of $x_{im}$ in PQN is:

\begin{equation} \label{2}
s_{im} = \sum_{k=1}^{K} \frac{e^{\alpha \langle x_{im}, C_{mk}\rangle}}{\sum_{j=1}^K e^{\alpha \langle x_{im}, C_{mj} \rangle}} C_{mk} =  \sum_{k=1}^{K} u_{mk} * C_{mk}
\end{equation}
where $\alpha$ is a scaling factor. When $\alpha \rightarrow +\infty$, $u_{mk} \rightarrow \mathbbm{1}(k=k_{*})$, which is a one-hot encoding vector with one in the $k_{*}$-th entry and zeros elsewhere. Here, $k_{*}=\argmax_{k}x_{im}^TC_{mk}$, represents the index of the most similar codewords for hard quantization. PQN avoids an infeasible derivative caused by $\argmax(\cdot)$ and allows the network to be optimized by a standard gradient descent algorithm. Based on PQN, researchers in~\cite{yu2020product} further develop RPQN and TPQN, which achieved a higher accuracy for image retrieval and accelerated video retrieval, respectively. More recently, a multiple exemplars learning (MLE) approach is proposed in~\cite{yu2021multiple}. Instead of learning a codebook shared by different classes, MLE learns a class-specific codebook consisting of multiple exemplars to partition the class-specific feature space. It makes the samples of different classes disentangled and improves retrieval accuracy.

One major problem when employing existing deep quantization methods to face image retrieval tasks is that the prior learning metrics are not sufficiently competent. 
For example, PQN~\cite{yu2020product} used an asymmetric triplet loss as the similarity metric and required a complex hard sample mining strategy during training. Moreover, the computational cost is prohibitive when using a large number of triplet samples. Therefore, it cannot learn the full dataset structure of the fine-grained face images. DPQ~\cite{klein2019end} applied a joint central loss based on classical softmax loss. Nevertheless, prior works on deep face recognition~\cite{wang2018cosface,deng2019arcface} show that angular margin-based methods have a superior discriminative ability. To provide sufficient discriminative power for PQ-based face image retrieval, this paper proposed a subspace-wise joint classification loss that maximizes discriminability for both the quantized and original features in each subspace. Another problem is the kind of codewords preferred in deep quantization models for better performance. Instead of learning codewords end-to-end that have significant variations in pairwise distance, we use predefined orthonormal codewords which have fixed 90-degree angular separation in between.

\section{Orthonormal Product Quantization} \label{sec:3}
\begin{figure}[h]
	\centering
	\includegraphics[width = 0.95\textwidth]{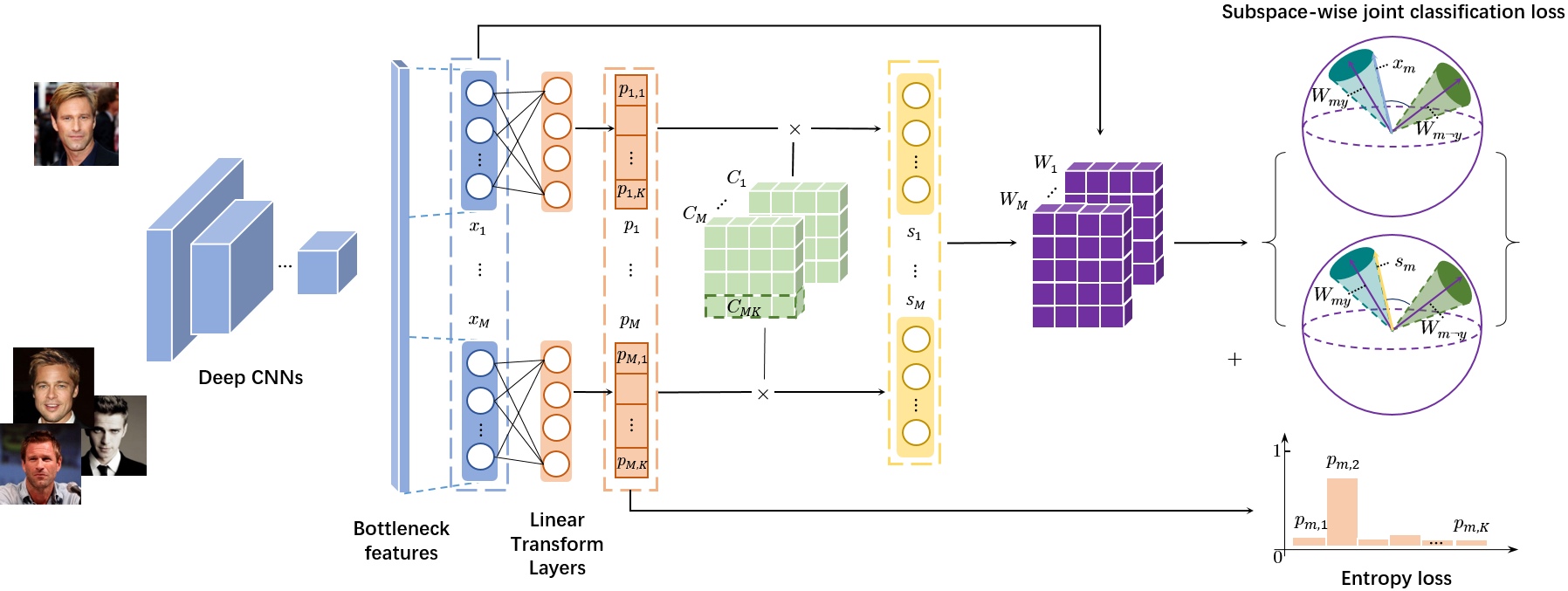}
	\caption{The overall training procedure of the OPQN method: $x_m$ represents the sub-vector from the bottleneck features. $x_{m}$ is projected by a linear transform layer followed by a softmax operation to produce the probability vector $p_{m}$. The soft quantization $s_{m}$ is constructed as a convex combination between $p_{m}$ and the orthonormal codewords $C_{m}$. $W_{m}$ represents the weight matrix of the subspace-wise classification loss. The complete loss function consists of classification loss and entropy loss. 
	}
	\label{opqn}
\end{figure}

We propose OPQN, a deep quantization-based method specialized for the face image retrieval task. The overview of the training procedure for OPQN is illustrated in Fig.~\ref{opqn}. We investigate the functionality of the distribution of codewords in the quantization process, then propose to use sets of predefined orthonormal codewords for quantization. The sub-vectors of deep features are transformed into assignment probabilities via feature-probability decorrelation. Then, they are combined with the predefined codewords to construct the soft quantizations. To provide sufficient discriminative power for PQ-based similarity search, a subspace-wise joint classification loss for the original sub-vectors and their soft quantizations is proposed. Besides, we impose an entropy-based regularization that allows for more precise quantization.

Here are the notations used in this paper. We denote a dataset with $N$ face images as $\{I_{i}\}_{i=1}^N$, and the corresponding label vector as $y\in \mathbb{R}^N$. For an input image $I_i$, $x_i \in \mathbb{R}^D$ is the bottleneck features as shown in Fig.~\ref{opqn} produced by the backbone network $f(\Theta)$, where $\Theta$ are the network parameters. $x_{i}$ is divided into $M$ disjoint sub-vectors with dimension $d={D}/{M}$, i.e., $x_i = [x_{i1}, x_{i2},\cdots,x_{iM}]$ where $x_{im} \in \mathbb{R}^{d}$. Assume the codebooks $C=[C_1, C_2,\cdots,C_M] \in \mathbb{R}^{M \times d \times K}$ and each codebook consists of $K$ codewords, i.e., $C_m = [C_{m1}, C_{m2},\cdots, C_{mK}] \in \mathbb{R}^{d \times K}$. The soft and hard quantizations of $x_{im}$ are $s_{im}$ and $h_{im}$, respectively. Concretely, $h_{im}=C_{mk_{*}}$, is the approximation of $x_{im}$ by the codeword with index $k_{*}$ in $m$-th codebook. Thus, the soft and hard quantizations of $I_i$ can be represented as $s_i = \{s_{im}\}_{m=1}^M$ and $h_i = \{h_{im}\}_{m=1}^M$, respectively. 

\subsection{Soft Quantization via Feature-Probability Decorrelation} \label{subsec:a}
The soft quantization shown in Eq.~(\ref{2}) implicitly encodes the similarity between sub-vectors and codewords. Consequently, how each codeword constitutes $s_{im}$, and how far the distance between $s_{im}$ and $C_{mk_{*}}$ is, are both difficult to observe. As the scaling factor cannot be set to positive infinity, there is always a gap between $s_{im}$ and $h_{im}$. Alternatively, OPQN learns the codewords assignment explicitly via an intermediate FC layer. Note that since the codewords are now predefined, it offsets the number of parameters in the intermediate layer. Thus, the total number of learnable parameters does not grow. Inspired by DPQ~\cite{klein2019end}, a linear transformation layer is built on top of each sub-vector $x_{im}$ individually. Denote the parameter matrices in all linear transform layers as $F=[F_{1}, F_{2},\cdots,F_{M}] \in \mathbb{R}^{M \times d \times K}$. For simplicity, we omit the biases in each layer. By appending a softmax function to layer outputs, we can formulate the probability of assigning the codeword $C_{mk}$ to the subvector ${x_{im}}$ as:

\begin{equation} \label{3}
    p_{im,k} = \frac{e^{x_{im}^T F_{mk}}}{\sum_{j=1}^{K}e^{x_{im}^T F_{mj}}}
\end{equation}
where $F_{mj}$ is the $j$-th column of the parameter matrix $F_m$. The $K$ probabilities are concatenated to the vector: $p_{im}= [p_{im, 1}, p_{im,2},\cdots,p_{im,K}] \in \mathbb{R}^{K}$. The soft quantization $s_{im}$ of $x_{im}$ in the proposed method is represented as: 

\begin{equation} \label{4}
    s_{im}= \sum_{k=1}^{K} p_{im,k} * C_{mk}
\end{equation}
Eq.~(\ref{4}) means each soft quantization $s_{im}$ is the convex combination of $\{C_{mk}\}$ with softmax coefficients $p_{im}$. Based on this, one can naturally derive the hard quantization as $h_{im}=C_{mk_{*}}$. Here, $k_{*}$ is the index of the codeword with the largest value in $p_{m}^i$, formulated as:
\begin{equation} \label{5}
    k_* = \argmax_{k} p_{im,k} \quad s.t. \quad k =1, 2,\cdots K
\end{equation}
Note that the proposed method does not compute the hard quantization in the training process, which avoids the problem of calculating the derivative of $\argmax(\cdot)$. Eq.~(\ref{5}) serves to encode database items during the testing query phase, which will be detailed in Section~\ref{adc}.

\subsection{Orthonormal Codewords Generation}
The above soft quantization method connects the sub-vectors and codewords by learning a linear transform matrix to represent the quantization composition explicitly. Thus, it is feasible to use predefined codewords for quantization. Assume there is a codebook $C_m$, with each column being a codeword. For any pair of codewords $C_{mi}$ and $C_{mj}$: $ 0 \leq \angle (C_{mi}, C_{mj}) \leq \pi$. The basic idea of codeword design is to improve the informativeness of quantization and reduce the redundancy in the codewords. We require each pair of codewords to have a sufficiently large angle between them and the variance of angles to be as small as possible. To eliminate side effects caused by the magnitudes, $C_{mi}$ and $C_{mj}$ should be normalized to unit length. In terms of these requirements, we apply the orthonormal vectors as codewords, which possess the desirable characteristics: $\|C_{mk}\|=1$ and $C_m^TC_m = I_K$. The orthonormal vectors are naturally of the unit norm and every two different orthonormal vectors keep a $\pi/2$ angular separation from each other. 

There are some other choices to generate a set of orthonormal vectors. For example, one can perform Singular Value Decomposition (SVD) on a random matrix and return the columns of the left-singular vectors as codewords. However, a better solution is to use deterministic orthonormal vectors that exclude the randomness biases caused by codewords themselves. Thereby, we utilize the cosine basis of Discrete Cosine Transform (DCT)~\cite{ahmed1974discrete}, in the DCT-\RNum{2} algorithm. Since DCT-\RNum{2} is defined by a set of orthogonal/orthonormal cosine basis functions, one can always obtain the exact same orthonormal vectors if only the dimension is specified. Suppose the dimension of sub-vectors, as well as codewords, are $d$, the basis matrix $A \in \mathbb{R}^{d \times d}$ in the DCT-\RNum{2} transform can be calculated as:
\begin{equation}
    A_{ij} = \cos{[\frac{j \pi}{d}(i+\frac{1}{2})]} \quad s.t. \; i, j=0,1,2,\cdots d-1
    \label{21}
\end{equation}

The procedure to generate orthonormal codewords using the cosine basis is summarized in Algorithm~\ref{alg:algorithm}. By several processing steps (steps 2 and 3) on $A$, we could obtain an orthogonal matrix $A^{\dagger}$, whose first $K$ column vectors are the desired codewords in one codebook. The orthonormal vectors multiplied by an orthogonal matrix are still orthonormal vectors. Thus, we iteratively multiply the previous codebook by $A^{\dagger}$ to get the new codebook, which guarantees the diversity between different sets of codebooks while still retaining the orthogonality of each codebook. Since the orthogonal matrices $A^{\dagger}$ are square, OPQN requires that the number of codewords is no more than the dimension of the sub-vectors. For a network with a bottleneck of 2048-dimensional features, OPQN can generate binary codes up to 64-bit ($8 \log_2 256$) in length, which is sufficient to cover most cases. 

{\linespread{1.3}
\begin{algorithm}[h]
\caption{Generation of deterministic orthonormal codewords}
\label{alg:algorithm}
\textbf{Input}: Features dimension $D$, number of codebooks $M$, number of codewords per codebook $K$, sub-features dimension $d$ ($d= D/M$ and $d \geq K$)
\\
\textbf{Output}: Codebooks $C\in \mathbb{R}^{M \times d \times K}$ 

\begin{algorithmic}[1]
\State Compute the cosine basis matrix $A$ according to Eq.~(\ref{21})
\State $A[:, 0] \leftarrow A[:, 0] / \sqrt{2}$ 
\State $A^\dagger \leftarrow \sqrt{2}A /\sqrt{d}$
\State $C_1 = A^\dagger[:, :K]$
\For {$m = 2 : M+1$}
    \State $C_m = A^\dagger * C_{m-1}$
\EndFor
\end{algorithmic}
\end{algorithm}
}

\subsection{Subspace-Wise Joint Classification Loss}
By substituting codewords in Eq.~(\ref{4}) with orthonormal codewords generated by Algorithm~\ref{alg:algorithm}, we can obtain the soft quantization of feature vectors in each subspace. These quantized features will be fed into the carefully designed objective function supervised with label information for discriminative retrieval. Meanwhile, it is natural that the bottleneck features directly limit the quantization performance. The well-learned original features should benefit the embedding of identity-specific clues in quantized features. Therefore, we propose to preserve the discriminability in both the original and soft quantized features. From another view, since original features and their quantized versions fall into several disjoint subspaces in PQ-based methods, the associated full identity information breaks into different partitions. For better classification, we expect intra-identity features and their soft quantizations to be separable from those belonging to other identities in each subspace. Thus, a set of subspace-wise classifiers are learned individually for each segment of $x_{im}$ and $s_{im}$.

We denote a fully connected layer containing a set of weight matrices as $W=[W_1, W_2, \cdots, W_M] \in \mathbb{R}^{M \times d \times C}$, where $C$ is the number of identity classes in the datasets. $W_{mc}$ represents the $c$-th column vector of $W_{m}$ in the $m$-th subspace. We normalize $W_{mc}$: $W_{mc} \leftarrow W_{mc}/\|W_{mc}\|_2$ that is commonly used in deep face recognition. Correspondingly, $x_{im}$ is also $\ell_2$ normalized: $x_{im} \leftarrow x_{im}/\|x_{im}\|_2$. Thus, the cosine similarity between $x_{im}$ and $W_{mc}$ is directly implied from their inner product. Specifically, $\cos{\theta}_{y_i, x_{im}}=x_{im}^T W_{m y_i}$, where ${\theta}_{y_i, x_{im}}$ represents the angle between $x_{im}$ with label $y_i$ and its corresponding weight vector $W_{m y_i}$. Inspired by the popular and effective line of angular margin-based deep face recognition~\cite{wang2018cosface,deng2019arcface}, we add a cosine margin $u$ between $\cos{\theta}_{y_i, x_{im}}$ and $\cos{\theta}_{y_{\neg i}, x_{im}}$. The introduced margin helps to enhance the intra-identity compactness and inter-class discriminability of the original features in each subspace. By formulating the angular margin into the softmax classification loss and summing up the loss terms coming from all the $M$ segments, we obtain the loss function concerning $x$ and $W$ as: 
\begin{equation} \label{7}
     L_{x} = \sum_{i=1}^N \sum_{m=1}^{M}-\log \frac{e^{r(\cos{\theta}_{y_i, {x_{im}}} - u)}}{e^{r(\cos{\theta}_{y_i,{x_{im}}} - u)} + \sum_{j \neq y_i} e^{r\cos{\theta}_{j, x_{im}}}} 
\end{equation}
where $r$ is a scaling factor for the normalized sub-features. We also $\ell_2$ normalize $s_{im}$ to remove the variation in the radius. Then, the cosine distance between $s_{im}$ and $W_{m}$ is indicated by their multiplication, and the angle between $s_{im}$ and $W_{m{y_i}}$ is denoted as $\theta_{y_i, s_{im}}$. Similarly, the angular margin-based loss function w.r.t. $s$ and $W$ is formulated as:
\begin{equation} \label{8}
     L_{s} = \sum_{i=1}^N \sum_{m=1}^{M}-\log \frac{e^{r(\cos{\theta}_{y_i, s_{im}} - u)}}{e^{r(\cos{\theta}_{y_i, s_{im}} - u)} + \sum_{j \neq y_i} e^{r\cos{\theta}_{j, s_{im}}}} 
\end{equation}
The values of margin $u$ and scaling factor $r$ in Eq.~(\ref{8}) are the same as in Eq.~(\ref{7}) to encourage consistency between $x_{im}$ and $s_{im}$. Combining Eq.~(\ref{7}) and Eq.~(\ref{8}), the joint similarity-preserving loss is represented as:
\begin{equation} \label{9}
     \min_{\Theta, F, W} L_{clf} = \frac{1}{2MN} \left( L_{x} + L_{s}\right)
\end{equation}
Eq.~(\ref{9}) targets subspace-wise intra-identity variance minimization and inter-identity variance maximization for both the original and quantized features. 

\subsection{Entropy Minimization for One-Hot Codewords Assignment}
The joint classification loss utilized soft quantization $s_{im}$ without considering hard quantization $h_{im}$ in training. However, we would like to reduce the discrepancy between $s_{im}$ and its corresponding original version $h_{im}$. The probability vector $p_{im}$, which takes the role of codewords assignment, should be close to one-hot encoding. Therefore, we propose an entropy-based regularization term to force the sub-features to move towards a single codeword while pushing it apart from other codewords. The entropy-based loss is formulated as:
\begin{equation} \label{10}
    L_{ent} = -\frac{1}{MN}\sum_{i=1}^{N} \sum_{m=1}^{M} \sum_{k=1}^{K} p_{im,k}\log{p_{im,k}} 
\end{equation}
$p_{im,k}\log{p_{im,k}}$ has the minimum value 0 if and only if $p_{im,k}=0$ or $p_{im,k}=1$. Under the constraint of $\sum_k p_{im,k}=1$, the proposed entropy loss tends to shape the distribution of $p_{im}$ into a pattern with a single peak at one index with small values elsewhere. By adding the loss from all the $M$ probability vectors, it aims to reduce the discrepancy between $s_{i}$ and $h_{i}$ for more precise quantization. Integrating $L_{clf}$ and $L_{ent}$ to obtain the finalized loss function of OPQN:
\begin{equation} \label{11}
   L = L_{clf} + \lambda L_{ent}
\end{equation}
where $\lambda$ is a balance weight of the entropy loss.

\subsection{Learning and Optimization}
The proposed OPQN contains three sets of learnable parameters: backbone network parameters $\Theta$, linear transform layer parameters $F$ and the classification weight $W$. We adopt the mini-batch strategy and stochastic gradient descent (SGD) in training and all parameters can be learned by back-propagation (BP). Denote $x_{im}^T$ multiplied by $F_{m}$ shown in Eq.~(\ref{3}) as $g_{m}=[g_{m1}, g_{m2}, \cdots, g_{mK}]$. The gradients of $p_{im}$ w.r.t $g_{m}$ can be computed as:
\begin{equation} \label{12}
  \frac{\partial p_{im,k}}{\partial g_{mk}} = p_{im,k}(1-p_{im,k}); \: \frac{\partial p_{im,k}}{\partial g_{mj (j \neq k)}} = -p_{im,k}p_{im,j} 
\end{equation}
Thus, we can derive the gradient of the soft quantization $s_{im}$ w.r.t. $g_{mk}$ by BP:
\begin{equation} \label{13}
  \frac{\partial s_{im}}{\partial g_{mk}} = \left [\frac{\partial s_{im}}{\partial p_{im}} \right]^T \frac{\partial p_{im}}{\partial g_{mk}} = p_{im,k}(C_{mk}-s_{im})
\end{equation}

Similarly, since $\partial L_{ent} / \partial p_{im} =-(1+\log{p_{im}})$, we obtain the derivatives of $L_{ent}$ w.r.t. $g_{mk}$ using Eq.~(\ref{12}) as:
\begin{equation} \label{14}
  \frac{\partial L_{ent}}{\partial g_{mk}} = p_{im,k} \Big (\sum_{j=1}^{K} p_{im,j}\log{p_{im,j}} - \log{p_{im,k}} \Big)
\end{equation}

Combining Eq.~(\ref{13}) and Eq.~(\ref{14}) and applying BP, the derivative of $L$ regarding $F_{mk}$ is:
\begin{equation} \label{15}
   \frac{\partial L}{\partial F_{mk}} = \left [ \frac{1}{2}\left (\frac{\partial L_{s}}{\partial s_{im}} \right)^T \frac{\partial s_{im}}{\partial g_{mk}} + 
   \lambda \frac{\partial L_{ent}}{\partial g_{mk}}\right ] x_{im}
\end{equation}
Likewise, the derivative of $L$ regarding $W$ is calculated by:

\begin{equation} \label{16}
   \frac{\partial L}{\partial W_{mk}} = \frac{1}{2}\left (\frac{\partial L_{x}}{\partial W_{mk}} + 
   \frac{\partial L_{s}}{\partial W_{mk}}
   \right )
\end{equation}

The derivative of $L$ w.r.t. $x_{im}$ is:
\begin{equation} \label{17}
   \frac{\partial L}{\partial x_{im}} = \frac{1}{2} \frac{\partial L_{x}}{\partial x_{im}} + \frac{1}{2}\left (\frac{\partial L_{s}}{\partial g_{mk}} + \lambda 
   \frac{\partial L_{ent}}{\partial g_{mk}}
   \right ) F_{mk}
\end{equation}
The complete training procedure of OPQN is summarized in Algorithm~\ref{alg:algorithm2}. 

{\linespread{1.3}
\begin{algorithm}[htbp]
\caption{OPQN Training Procedure}
\label{alg:algorithm2}
\textbf{Input}: Training set $\{I_{i}\}_{i=1}^N$ with labels $y$, the network $f(\cdot)$, the dimension of codebooks: $M\times d \times K$;\\
\textbf{Initialization}: Backbone network parameters $\Theta$, linear transform layer $F$, classification weight matrix $W$;

\begin{algorithmic}[1]
\State Generate orthonormal codewords by Algorithm~\ref{alg:algorithm};
\Repeat
\State Randomly sample a mini-batch data from the training set;
\State \parbox[t]{310pt}{Feed forward the mini-batch images through the model and compute $x_{i}=f(\Theta;I_{i})$ for each image;\strut}
\State Calculate the objective function $L$ according to Eq.~(\ref{11});
\State \parbox[t]{310pt}{Compute the derivatives of $L$ w.r.t. $W$, $F$ and $x_{im}$ according to Eq.~(\ref{16}), Eq.~(\ref{15}) and Eq.~(\ref{17}), respectively;\strut}
\State \parbox[t]{310pt}{Back propagate the gradients to the backbone network, then update the parameters $W$, $F$ and $\Theta$;\strut}
\Until{Convergence}
\end{algorithmic}
\end{algorithm}
}

\subsection{Asymmetric Distance Comparison for Retrieval} \label{adc}
Following previous works~\cite{yu2018product,klein2019end,yu2020product,jang2020generalized}, we apply asymmetric quantization distance (AQD)~\cite{jegou2010product} as the similarity metric in the search phase. AQD enables using soft quantizations to represent a query but hard quantizations to encode database images. It has the advantages of both memory footprint reduction and retrieval speed acceleration. To this end, the query and database items in the search phase are processed with different procedures. 

Given a query image $q$, we propagate it through the model until the linear transform layers. These outputs are passed to the softmax function, as in Eq.~(\ref{3}), to obtain the probability vector $p_{qm}$ of each subvector $x_{qm}$. Then, combined with $C_{m}$ to obtain the soft quantization $s_{qm}$, as in Eq.~(\ref{4}). For each database image $I_{i}$, we pre-compute $\{p_{im}\}$ following the same procedure as the query image. $I_i$ is associated with its hard quantization $\{h_{im}\}$ via the indices of the largest values in $\{p_{im}\}$. Suppose that we have a matrix $B \in \mathbb{R}^{N \times M}$, then each element $b_{im}$ of $B$ stores the index $k_*$ of codewords in Eq.~(\ref{5}). Therefore, the Euclidean distance between $q$ and $I_i$, is computed as:
\begin{equation} \label{18}
     AQD(q, I_{i}) = \sum_{m=1}^M \|s_{qm} - h_{im}\|_2^2 = \sum_{m=1}^M \|C_m p_{qm} - C_{mb_{im}}\|_2^2
\end{equation}
Note the orthogonality of $C_{m}$. By expanding the right-hand side of Eq.~(\ref{18}) and eliminating the constant and the term irrelevant to $C_{mb_{im}}$, we can derive the following equation:

\begin{equation} \label{20}
    \argmin_{i} AQD(q, I_{i}) = \argmin_{i} \sum_{m=1}^M - 2{p_{qm}}^T C_m^T C_{mb_{im}} = \argmax_{i} \sum_{m=1}^M p_{qm,b_{im}}
\end{equation}

From Eq.~(\ref{20}), $\argmin_{i} AQD(q, I_{i})$ depends on $\{p_{qm}\}_{m=1}^M$ and $\{b_{im}\}_{m=1}^M$. It indicates the quantization similarity comparison between the query and any database item can be realized efficiently by indexing $p_{qm}$ with LUTs.
Specifically, we build $M$ LUTs, denoted as $\{LUT_m\}$, w.r.t. $M$ probability vectors $\{p_{q}\}$, where $LUT_m[i] = b_{im}$. Since the matrix $B$ can be pre-computed, it only takes several addition calculations in the searching process. Compared with DQN~\cite{cao2016deep} and DPQ~\cite{klein2019end}, OPQN does not require explicit online reconstruction of the soft quantization or calculation of the Euclidean distance between the soft quantization and each codeword. Therefore, it is more scalable and time-efficient to handle a query that arrives on-the-fly. Besides, the codebooks in our method are data-independent and predefined. This allows models to use the same codebook for different datasets, helping to lower the system's storage cost. The retrieval procedure of OPQN is summarized in Algorithm~\ref{alg:algorithm3}.

{\linespread{1.3}
\begin{algorithm}[htbp]
\caption{OPQN Top-$k$ Retrieval Procedure}
\label{alg:algorithm3}
\textbf{Input}: Database images $DB=\{db_i\}_{i=1}^{\mid DB\mid}$, query set images $Q=\{q_i\}_{i=1}^{\mid Q \mid}$, the trained model;\\
\textbf{Output}: Top $k$ instances in $DB$ for each $q_i$;
\begin{algorithmic}[1]
\State Forward pass $DB$ through the model in advance, and pre-compute the indices matrix $B$ according to Eq.~(\ref{3}) and Eq.~(\ref{5});
\State LUTs construction for $DB$ based on matrix $B$;
\For {$i$ in $1,2,\cdots, \mid Q \mid$:}
\State Forward pass $q_{i}$ through the model and compute $p_{q_{i}m}$ by Eq.~(\ref{3});
\State \parbox[t]{310pt}{Compute similarity between $q_i$ and each database image by Eq.~(\ref{20}) using LUTs, and sort the results in descending order;\strut}
\EndFor
\end{algorithmic}
\end{algorithm}
}

\section{Experiments and Results} \label{sec:4}
\subsection{Datasets and Evaluation Metrics} \label{datasets}
To demonstrate the performance of the proposed OPQN, we conduct experiments on four commonly-used publicly available datasets: FaceScrub~\cite{ng2014data}, CFW-60K~\cite{li2015two}, VGGFace2~\cite{cao2018vggface2}, and YouTube Faces~\cite{wolf2011face}. For seen identity retrieval, the training-testing split is used as database images and queries, respectively, for each dataset. For simplicity, we refer to the evaluation of seen identities as the standard retrieval. 
Three kinds of evaluation metrics are applied to evaluate the quality of the retrieval: mean average precision (MAP), precision-recall (PR) curve, and precision w.r.t. top $T$ returned images (P@$T$). The details of each dataset and their corresponding protocols are:

\textbf{FaceScrub}~\cite{ng2014data} contains 106,863 face images of 530 celebrities with about 200 images per identity. We use the same training-testing split as in~\cite{lin2017discriminative, tang2018supervised,tang2018discriminative}: five images per identity are selected for testing, and the remaining images are used for training. All the face images have been cropped and resized to 32$\times$32.

\textbf{CFW-60K}~\cite{li2015two} is a dataset containing 60,000 images of 500 identities. As in~\cite{zhang2021deep}, we use the official test split in CFW-60K, which has 10 images per identity and a total of 5000 images for testing. Among the other images, a total number of 55,000 images with category labels were used for training. All the face images have the same size of 32$\times$32. 

The \textbf{VGGFace2}~\cite{cao2018vggface2} dataset contains 3.31 million images of 9,131 identities, officially split into 8,631 identities for training and 500 identities for testing. The identities in the training testing split are disjoint. We choose 2,787 identities, with approximately 300 images each, from the official training set. 50 images per identity were taken for testing, with the rest used for training. 
In the unseen identity retrieval protocol, we take all the identities in the official testing set. 50 images per identity are used as queries, with the remainder used as a database. All the images in VGGFace2 are cropped and aligned following the instructions of MTCNN~\cite{zhang2016joint}. Each image is resized to 112$\times$112.

\textbf{YouTube Faces}~\cite{wolf2011face} contains thousands of videos of 1,595 celebrities. 
Following the configuration in~\cite{lin2017discriminative}, we select 40 images per identity for training and 5 images for testing. The employed dataset has 63,800 training images and 7,975 testing images. Each image is resized to 32$\times$32. 

\subsection{Experiment Settings}
We compare OPQN with a series of hashing-based and PQ-based methods. The hashing methods include DDQH~\cite{tang2018discriminative}, DCWH~\cite{zhe2019deep}, CSQ~\cite{Yuan_2020_CVPR}, DCDH~\cite{zhang2021deep}, and DPAH~\cite{wang2020deep}. 
Since DDQH and DCDH in their papers apply different backbone networks from OPQN, we denote the original results as DDQH and DCDH, while our unified implementations as DDQH* and DCDH*, respectively. For PQ-based methods, we compare OPQN with DPQ~\cite{klein2019end} and GPQ~\cite{jang2020generalized}, which are initially designed for general image retrieval. We also present results of the variant of OPQN, namely OPQN-l2q, which employs the same learning metric as OPQN but learns codewords instead of predetermining them.

To ensure a fair comparison, we unify the backbone network for all the binary hashing and PQ-based methods. Thus, the difference in the networks between binary hashing and PQ-based methods only exists in the last few layers, which distinguish the approaches from each other. Without loss of generality, the employed backbone network is based on the ResNet20~\cite{he2016deep} architecture. \footnote{The applied network architecture is illustrated in the supplementary material.}
Similar architectures have been used in prior deep face recognition works~\cite{wang2018cosface,deng2019arcface}. For PQ-based methods, the outputs of the last convolutional layer are flattened and projected into the FC1 layer to generate the bottleneck features shown in Fig.~\ref{opqn}. In contrast, for hashing-based methods, the flattened convolutional layer's outputs are first transformed into 512 dimensions by the FC1 layer, followed by an FC hashing layer to produce hashing outputs with the expected code length. Batch normalization~\cite{ioffe2015batch} is used after each FC layer in both methods.

\subsection{Implementation Details}
We evaluate the performance of OPQN under code lengths ranging from 16 to 64 bits. During training, OPQN uses a mini-batch SGD algorithm for optimization with momentum of 0.9 and weight decay of 5e-4. For small datasets, i.e., FaceScrub~\cite{ng2014data} and CFW-60K~\cite{li2015two}, the initial learning rate is set to 0.1 and decayed by 0.5 every 35 epochs, while in VGGFace2~\cite{cao2018vggface2} dataset,
the initial learning rate is set to 0.01 and decayed by 0.5 every 20 epochs. The batch size is fixed to 256 for all datasets, and the whole network is trained for 200 epochs. From cross-validation, the parameter settings for OPQN are the scaling factor $r=40$, the margin $u=0.4$, and the balance weight $\lambda=0.1$.
We apply the same data augmentation process including random cropping and random horizontal flipping to all methods during training. Experiments of the compared methods use the codes available from the original authors if possible. Otherwise, we carefully implement the methods. All experiments are performed on two Nvidia RTX-2080 GPU cards and implemented with PyTorch. The source codes are released at \url{https://github.com/mzhang367/opqn}.

\subsection{Seen Identity Retrieval}
We first conduct experiments on the FaceScrub and CFW-60K datasets, following the standard retrieval protocol in prior works. We evaluate the binary codes under 16, 24, 36, and 48 bits, with the number of codebooks, $M$, set empirically to 2, 4, 6, and 8, respectively. Thus, the number of codewords in each codebook, $K$, are accordingly 256, 64, 64, and 64. Note that the bottleneck features, with dimension $D$, should fulfill the conditions $D/M \leq K$ and $D \mid M$. For simplicity, we fix $D$ to 512 for all cases except 36 bits where $D$ is set to 516 to be divisible by the number of codebooks, i.e., 6. The other PQ-based methods adopt the same codebook and codeword settings as the proposed OPQN. 

The MAP results on FaceScrub and CFW-60K datasets are summarized in Table~\ref{table:map_single}. OPQN outperforms all the other methods over all code lengths. OPQN achieves average performance improvements of 3.83$\%$ and 2.26$\%$ over the state-of-the-art deep hashing method DCDH* on the FaceScrub and CFW-60K datasets, respectively. The superiority of OPQN is more prominent under short code lengths. For example, under 16-bit codes, it outperforms the second place DCDH* by a margin of 5.36$\%$ and 4.92$\%$ on two datasets, respectively. 

{\linespread{1.2}
\begin{table}[htbp]
    \small
    \centering
    \caption{MAP ($\%$) results on FaceScrub, CFW-60K, and VGGFace2 datasets under the standard retrieval setting}
    \setlength{\tabcolsep}{0.5mm}{
    \begin{tabu}{ccccccccccccc}
        \toprule
        \multirow{2}{*}{Method} & \multicolumn{4}{c}{FaceScrub} & \multicolumn{4}{c}{CFW-60K} & \multicolumn{4}{c}{VGGFace2}\\
        \cmidrule(lr){2-5}\cmidrule(lr){6-9} \cmidrule(lr){10-13}
        &16-bit &24-bit &36-bit &48-bit &16-bit &24-bit &36-bit &48-bit &24-bit &36-bit &48-bit &64-bit\\
        \midrule
        DDQH    &-  &44.82  &50.71  &51.91  &-  &-  &-  &-  &-  &-  &-  &- \\
        DDQH*  &83.93  &85.61  &87.14  &88.28 &78.32  &78.80  &80.69  &82.22 &81.19 &90.14 &91.67 &92.89 \\
        DCWH &83.52  &84.45  &85.62  &88.72  &70.14 &72.51  &74.80  &78.39 &34.23 &52.24 &64.58 &72.59 \\
        DPAH &83.98  &88.59  &90.06  &90.41  &76.02 &82.66  &83.99  &84.54 &81.59 &87.15 &88.97 &90.42 \\
        CSQ &71.23  &80.32  &82.06  &86.61  &69.69 &72.19  &78.44  &83.35 &70.72 &79.34 &83.04 &83.71 \\
        DPQ &38.70  &84.05  &90.43  &90.71  &30.43 &57.61  &70.37  &70.55 &71.05 &77.03  &84.35  &86.81 \\
        GPQ &63.38  &80.36  &85.93  &86.99  &64.02 &49.31  &61.46  &71.90 &67.90 &70.41 &72.86 &-\\
        DCDH  &-  &77.79  &83.47  &84.64  &-  &81.68  &83.56  &85.48 &-  &-  &-  &- \\        
        DCDH* &84.96 &87.18 &89.53 &91.43 &80.55 &86.08 &86.69 &87.16 &87.52 &90.95 &91.81 &92.32 \\
        \midrule
        OPQN (Ours)  &$\mathbf{90.32}$  &$\mathbf{91.54}$  &$\mathbf{92.70}$  &$\mathbf{93.85}$ &$\mathbf{85.47}$  &$\mathbf{86.37}$  &$\mathbf{88.26}$  &$\mathbf{89.40}$ &$\mathbf{89.86}$  &$\mathbf{95.08}$  &$\mathbf{95.04}$  &$\mathbf{95.29}$ \\
        OPQN-l2q &62.73  &89.71  &91.62  &92.54  &52.78 &83.51  &87.25  &87.80 &76.00 &85.75  &91.27  &90.59 \\
        \bottomrule
    \end{tabu}}
    \label{table:map_single}
\end{table}
}

The other deep quantization models, DPQ and GPQ, do not perform as well as deep hashing-based methods. Particularly, the performances of DPQ and GPQ are poor for 16-bit codes. This may be due to their applied learning metrics being insufficient to extract discriminative features given that the complete visual information attached to the feature vector is divided into several subspaces. 
It is worth noting that the variant, OPQN-l2q, performs worse than OPQN, with relatively minor decreases for 24 to 48-bit codes but is 27.59$\%$ and 32.69$\%$ lower for the 16-bit codes on FaceScrub and CFW-60K, respectively. The results strongly support the effectiveness of the proposed orthonormal scheme. As shown in Fig.~\ref{ang_dst}, we compare the angular distribution of pairwise codewords of four methods on FaceScrub under 16-bit codes. Fig.~\ref{ang_dst} implies that the orthonormal codewords contribute to better preserving the visual information in the subspace, by which the codewords are evenly distributed with a specific and moderate (90 degrees) separation from each other. 

We further conduct experiments on the employed VGGFace2 dataset under four code lengths, 24, 36, 48, and 64 bits. The number of codebooks is set to 3, 4, 6, and 8, respectively. Accordingly, the number of codewords in each codebook is 256 except for the 36-bit case where it is 512. The dimension $D$ of the bottleneck features in OPQN is set to $D=MK$. For a fair comparison, the DPQ~\cite{klein2019end} method is evaluated with the same configuration of $D$, $M$, $K$, while the GPQ~\cite{jang2020generalized} method, could only perform well with a large value of $K$ in this case. Thus, we fix $K$ as 4096 ($2^{12}$), and $M$ as 2, 3, and 4 to obtain 24-bit, 36-bit, and 48-bit codes, respectively. We adopt $D=2048$ for 24 and 48-bit codes, and 2049 for 36-bit codes. The 64-bit result for GPQ is not shown as it does not support a reasonably large value of $K$.

The MAP performance on the VGGFace2 is shown in the right part of Table~\ref{table:map_single}. It is clear that OPQN outperforms all the other methods over all code lengths, exceeding the two best competitors, DCDH~\cite{zhang2021deep} and DDQH~\cite{tang2018discriminative}, by 3.18$\%$ and 4.85$\%$ on average, respectively. At 36 bits, OPQN's more than 95$\%$ MAP is 4.13$\%$ higher than the second place DCDH*. However, OPQN-l2q and DPQ have much poorer performance than their hashing-based counterparts, DCDH* and DDQH*, especially at 24 bits. 

\begin{figure}[htbp]
	\centering
	\includegraphics[width = 1.0\textwidth]{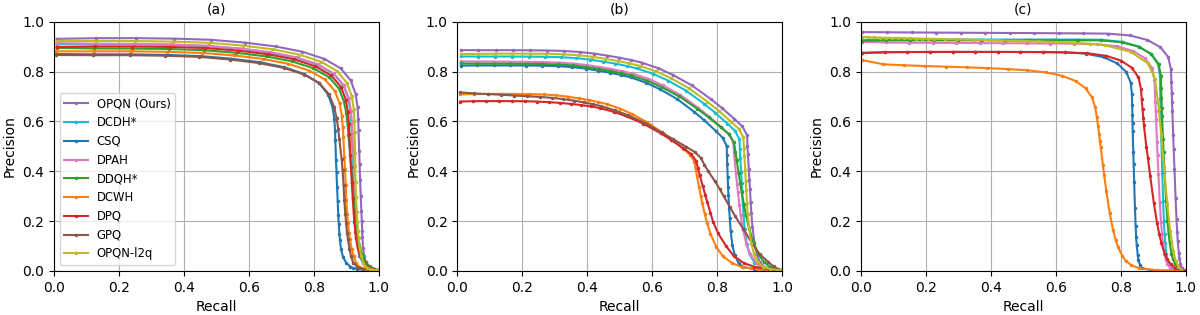}
	\caption{Performance measured by PR-curves: (a) 48-bit codes on FaceScrub, (b) 48-bit codes on CFW-60K, (c) 64-bit codes on VGGFace2.}
	\label{pr_three}
\end{figure}

\begin{figure}[!hbp]
	\centering
	\includegraphics[width = 1.0\textwidth]{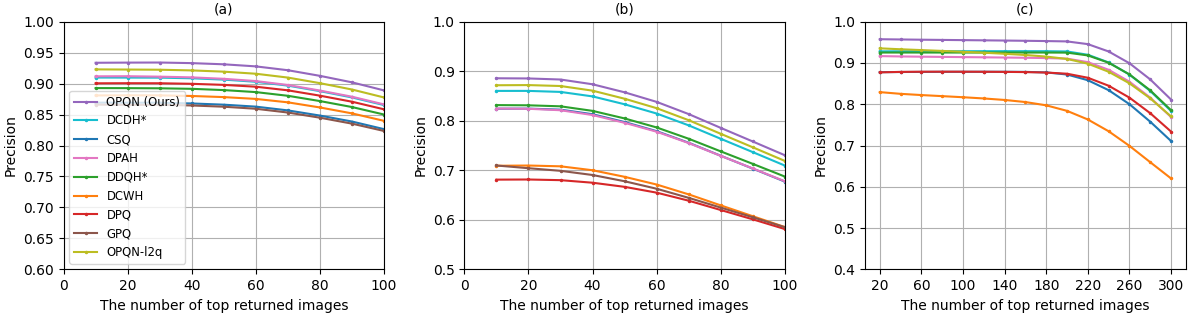}
	\caption{Performance w.r.t. different P@$T$ on three datasets: (a) 48-bit codes on FaceScrub, (b) 48-bit codes on CFW-60K, (c) 64-bit codes on VGGFace2.}
	\label{topk_three}
\end{figure}
We then evaluate the performance on the PR-curve as shown in Fig.~\ref{pr_three}.\footnote{More PR-curve results are given in the supplementary material.} We can see the PR-curves of OPQN almost always span outermost from the top-left to the bottom-right corner of the whole figures for three datasets. This means that OPQN can maintain a higher precision with an increase in the recall score. Since face image retrieval system users generally only look at top-ranking images, it is also necessary to evaluate the retrieval accuracy in terms of different numbers of top-returned images. Thus, Fig.~\ref{topk_three} plots P@$T$ curves on the three datasets. One can see that OPQN always provides superior precision scores over a moderate quantity of returns. Specifically, OPQN is the only method that maintains higher than 80$\%$ precision in the top 300 positions in VGGFace2, which has an average number of 300 images per identity in the database. Considering the challenging variation in the VGGFace2, retrieving more than 80$\%$ of related images is satisfying. Combing MAP and PR-curves results, we know that OPQN is robust to different evaluation metrics. 

\begin{figure}[!hbp]
	\centering
	\includegraphics[width = 0.50\textwidth]{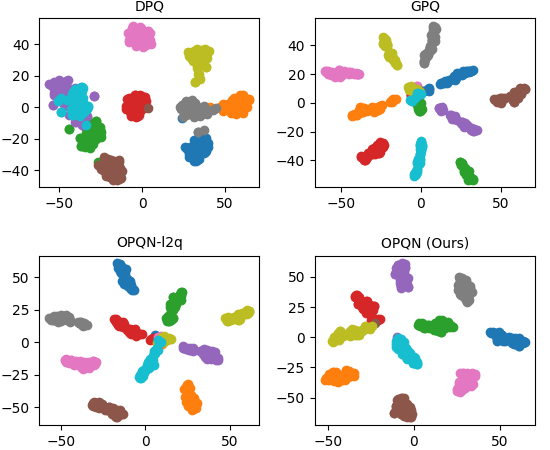}
	\caption{Visualization of 10-class deep feature representations in a subspace produced by DPQ, GPQ, OPQN-l2q, and OPQN. Each color represents a unique identity.}
	\label{vis_tsne}
\end{figure}

Finally, to intuitively compare the feature representations in a subspace learned by different deep quantization methods, we illustrate visualizations of the sub-vectors under 36-bit codes on the VGGFace2 dataset in Fig.~\ref{vis_tsne}. Ten identities are randomly selected from the testing set, and the trained deep quantization models are deployed on the samples to generate feature vectors directly. Instead of using the complete vector for visualization, we visualize sub-vectors, for which the quantization is performed by each codebook individually. In the experiment, the first 512-dimensional features are split from the 2048-dimensional bottleneck features. t-SNE~\cite{maaten2008visualizing} is applied to map the high-dimensional features of each method to 2-dimensional. From Fig.~\ref{vis_tsne}, OPQN produces the most separable feature representations in the subspace with the least overlapping when compared with other methods. 

To summarize experiment results under the standard retrieval setting, the superiority of OPQN is due to the following reasons. 1) The orthonormal codewords help enhance the informativeness of quantization and reduce redundancy in the codewords. The soft quantization strategy explicitly decorrelates the feature and probability, providing greater flexibility in the composition of the quantization when codewords are predefined. 2) OPQN can maximize the discriminability in soft quantizations, and in the original features, by benefiting from the joint subspace-wise classification loss, while the optimization of the original features facilitates obtaining better quantization representations. 

\subsection{Unseen Identity Retrieval} \label{cross-domain}
Different from the standard retrieval setting, where the evaluation dataset contains the same set of classes as the training dataset, unseen identity retrieval~\cite{klein2019end,sablayrolles2017should} used a set of unseen classes as queries and the database in the query stage. 
The employed protocol is given as follows. The deep hashing/quantization models have been pre-trained on the VGGFace2 training set for standard retrieval. These models are used for feature extraction and representation of unseen identity images. We conduct experiments on the database-query split of three datasets, i.e., VGGFace2 official testing set, CFW-60K, and YouTube Faces datasets, as described in Section~\ref{datasets}. In addition to MAP, P@$T$ is used to emphasize the top ranking. Specifically, P@10 is used for the VGGFace2, while P@5 is used for CFW-60K and YouTube Faces datasets. 

{\linespread{1.2}
\begin{table}[!hbp]
    \centering
    \small
    \caption{MAP ($\%$) results on VGGFace2, CFW-60K and YouTube Faces under the unseen identity retrieval setting}
    \setlength{\tabcolsep}{0.5mm}{
    \begin{tabular}{ccccccccccccc}
        \toprule
        \multirow{2}{*}{Method} & \multicolumn{4}{c}{VGGFace2} & \multicolumn{4}{c}{CFW-60K}  & \multicolumn{4}{c}{YouTube Faces}\\
        \cmidrule(lr){2-5}\cmidrule(lr){6-9} \cmidrule(lr){10-13} 
        &24-bit &36-bit &48-bit &64-bit &24-bit &36-bit &48-bit &64-bit &24-bit &36-bit &48-bit &64-bit\\
        \midrule
        DDQH*  &7.71  &9.71  &12.39  &14.69  &6.52  &8.61  &9.87  &11.97 &4.53  &6.85 &9.00  &11.29 \\
        DCWH &2.68  &4.65  &6.48  &8.13  &2.08 &3.59  &4.98  &6.24 &2.71  &5.25  &7.80  &9.63\\
        DPAH &2.71  &4.99  &6.25  &11.22  &2.55 &4.04  &5.12 &8.68 &2.26  &3.86  &5.72  &9.43\\
        CSQ  &2.30  &2.86  &3.36  &3.82  &2.25 &2.39  &3.09  &3.70 &2.41 &2.75 &3.73 &4.35\\
        DCDH* &3.15 &5.89 &7.00 &9.05 &3.36 &5.70 &6.21 &7.35 &2.56  &4.33  &4.89  &6.40\\
        DPQ &7.67  &8.21  &8.99  &13.68  &5.51 &6.22  &6.82 &10.01 &5.20  &6.08  &7.02 &10.75\\
        GPQ &10.24  &10.69  &11.24  &-  &8.52 &8.96  &9.42  &- &4.24  &4.69  &5.03  &-\\
        \midrule
        OPQN (Ours)  &$\mathbf{15.29}$  &$\mathbf{22.27}$  &$\mathbf{25.93}$  &$\mathbf{34.19}$ &$\mathbf{12.81}$  &$\mathbf{18.08}$  &$\mathbf{20.75}$  &$\mathbf{25.54}$ &$\mathbf{10.44}$  &$\mathbf{16.12}$  &$\mathbf{20.79}$  &$\mathbf{26.59}$\\
        OPQN-l2q &7.29  &10.84  &15.50  &22.00  &6.06 &8.77  &12.45  &17.61 &6.21 &8.27  &16.00  &23.33\\
        \bottomrule
    \end{tabular}}
    \label{table:map_cross}
\end{table}
}

The MAP and P@$T$ results of unseen identity retrieval on three datasets are shown in Tables~\ref{table:map_cross} and~\ref{table:ptop_cross}, respectively. From Table~\ref{table:map_cross}, OPQN outperforms other methods by a distinct margin over all the compared code lengths. Its MAP performance surpasses the second place GPQ by 10.44$\%$ and 8.25$\%$ on average, respectively, on VGGFace2 and CFW-60K datasets. On YouTube Faces, it exceeds the second place DDQH* by 10.57$\%$ on average. Compared with the l2q variant, OPQN generalizes better with an average superiority of 10.51$\%$, 8.07$\%$, and 5.03$\%$ on VGGFace2, CFW-60K, and YouTube Faces, respectively. The results confirm the effectiveness of the proposed orthonormal constraint on codewords to both standard retrieval and unseen identity retrieval. 
{\linespread{1.2}
\begin{table}[!hbp]
    \centering
    \small
    \caption{P$@T$ ($\%$) results on VGGFace2, CFW-60K and YouTube Faces under the unseen identity retrieval setting}
    \setlength{\tabcolsep}{0.50mm}{
    \begin{tabular}{ccccccccccccc}
        \toprule
        \multirow{2}{*}{Method} & \multicolumn{4}{c}{VGGFace2 (P$@10$)} & \multicolumn{4}{c}{CFW-60K (P$@5$)}  & \multicolumn{4}{c}{YouTube Faces (P$@5$)}\\
        \cmidrule(lr){2-5}\cmidrule(lr){6-9} \cmidrule(lr){10-13} 
        &24-bit &36-bit &48-bit &64-bit &24-bit &36-bit &48-bit &64-bit &24-bit &36-bit &48-bit &64-bit\\
        \midrule
        DDQH*  &19.85  &27.28  &33.91  &40.00 &13.94  &20.14  &23.49  &28.44 &11.77  &17.48  &23.06  &28.05\\
        DCWH &13.67  &23.13  &30.58  &35.98  &8.46 &14.26  &20.24  &24.35 &10.01  &18.42  &25.77  &30.47\\
        DPAH &12.21  &21.61  &27.53  &40.38  &8.21 &14.29  &17.58  &26.73 &7.74  &13.43  &19.02  &28.13\\
        CSQ  &12.55  &16.15  &18.03  &22.02  &7.52 &9.14  &11.65  &12.97 &8.48 &10.12 &12.36 &14.91\\
        DCDH* &14.26 &20.63 &29.15 &34.89 &10.03 &15.83 &18.28 &23.36 &8.53  &13.21  &16.20  &20.42\\
        DPQ &21.51  &26.21  &34.07  &40.57  &14.92 &17.68  &19.00 &26.77 &12.83  &14.65  &16.00 &28.51\\
        GPQ &20.51  &21.65  &22.58  &-  &16.58 &16.82  &17.24  &- &8.17  &9.19  &9.48  &-\\
        \midrule
        OPQN (Ours)  &$\mathbf{39.87}$  &$\mathbf{51.31}$  &$\mathbf{58.23}$  &$\mathbf{69.32}$ &$\mathbf{30.50}$  &$\mathbf{40.90}$  &$\mathbf{45.00}$  &$\mathbf{56.28}$ &$\mathbf{22.17}$  &$\mathbf{35.56}$  &$\mathbf{45.77}$  &$\mathbf{55.05}$\\
        OPQN-l2q &26.38  &36.64  &47.88  &59.76  &18.94 &25.08  &36.91  &47.50 &17.97 &23.48  &35.19  &46.49\\
        \bottomrule
    \end{tabular}}
    \label{table:ptop_cross}
\end{table}
}

As for P@$T$ results in Table~\ref{table:ptop_cross}, the pre-trained OPQN model performs significantly better than other methods with nearly 70$\%$ P$@10$ and more than 55$\%$ P$@5$ results on VGGFace2 and other two datasets at 64 bits. And the results are 28.75$\%$, 27.84$\%$, and 24.58$\%$ higher than the second best methods DPQ (VGGFace2), DDQH (CFW-60K) and DCWH (YouTube Faces), respectively. The P@$T$ performance of OPQN averaged over all bit values on the VGGFace2, CFW-60K, and YouTube Faces datasets are better than the second best methods DPQ, DDQH, and DCWH by 24.09$\%$, 21.67$\%$ and 18.47$\%$, respectively. Once again, using predefined orthonormal codewords considerably improves the performance of OPQN over OPQN-l2q, inducing an average improvement of 12.02$\%$ (VGGFace2), 11.06$\%$ (CFW-60K), and 8.86$\%$ (YouTube Faces) on three datasets. Since retrieving all the related samples is much more challenging than retrieving the most related samples, one can find the substantial performance gap between MAP and P@$T$ results comparing Tables~\ref{table:map_cross} and~\ref{table:ptop_cross}. Considering the high priority of top returned items in a face image retrieval system, the long-bit P@$T$ results of OPQN are still encouraging. 

\begin{figure}[htbp]
	\centering
	\includegraphics[width = 0.5\textwidth]{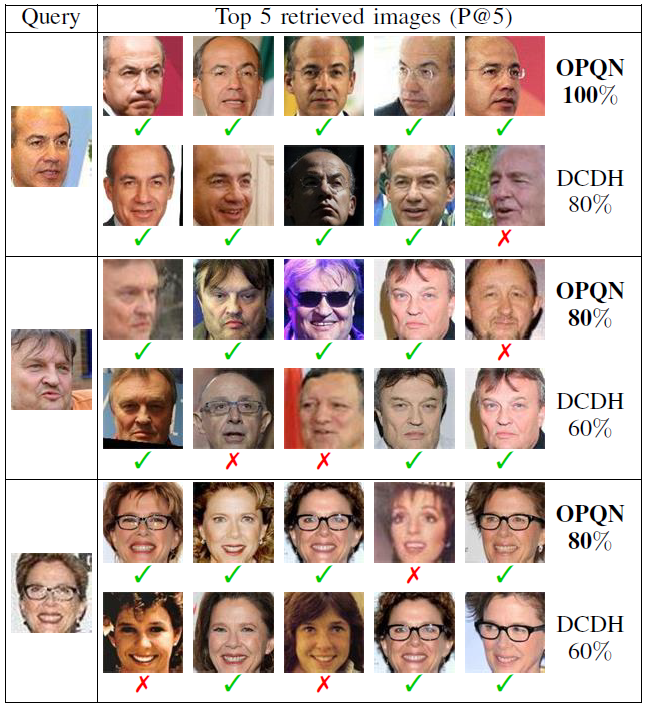}
	\caption{Examples of top-5 retrieved images by OPQN and DCDH under the unseen identity retrieval setting. The models are pre-trained with 64-bit codes and evaluated on the VGGFace2 dataset.}
	\label{exp_ret}
\end{figure}

There are two noteworthy observations from Tables~\ref{table:map_cross} and~\ref{table:ptop_cross}. Firstly, OPQN and OPQN-l2q, greatly outperform the baseline deep hashing methods, especially on long bits. However, OPQN-l2q is inferior to their deep hashing competitors in the task of seen identity retrieval (Table~\ref{table:map_single}). The reason is that PQ-based methods use real-valued codewords to reduce the deviations generated during encoding. With the exponential number of combinations on codewords, it can realize more fine-grained distance measurements between the database and queries. We further visualize the top samples returned by OPQN in comparison with DCDH, which is the second best method for standard retrieval, in Fig.~\ref{exp_ret}. It is straightforward to see that OPQN can return more truly relevant items when given an unseen identity as the query. The second observation is that the unseen identity retrieval performance, measured by both MAP and P$@T$, consistently improved with increasing code length for all methods. It indicates that longer binary code representations effectively improve the generalization ability of deep hashing/quantization methods. 

\section{Discussion}\label{sec:5}
\subsection{Ablation Study}
We investigate four variants of OPQN that apply different loss function designs, namely OPQN-A, OPQN-C, OPQN-S, and OPQN-W.

\textbf{OPQN-A}: In the OPQN-A variant, the learning metric only utilizes information within soft representations for training and the discriminability of the original features is discarded. In other words, the classification loss $L_{clf}$ in OPQN-A only contains the part derived in Eq.~(\ref{8}).  \textbf{OPQN-C}: The OPQN-C variant concatenates the composed soft representations and original sub-vectors in all the subspaces to full vectors before being fed into the classifier. 
Thus, it only utilizes the visual information of $x_i$ and $s_i$ in the concatenated space. \textbf{OPQN-S}: The OPQN-S variant replaces the angular margin-based classifier with a traditional softmax classifier as used in DPQ. Thus, no $\ell_2$ normalization is applied to sub-vectors, soft representations, or weight vectors. It still utilizes soft quantization and the original feature vectors in each subspace for training. \textbf{OPQN-W}: The OPQN-W variant represents the finalized objective function without the regularization term $L_{ent}$ shown in Eq.~(\ref{10}). This variant is set to observe to what extent the one-hot encoding of codewords can improve the performance of image retrieval.

{\linespread{1.2}
\begin{table}[htbp]
    \centering
    \small
    \caption{MAP ($\%$) results of three variants of OPQN for seen identity retrieval}
     \setlength{\tabcolsep}{1.0mm}{
    \begin{tabular}{ccccccccc}
        \toprule
        \multirow{2}{*}{Method} & \multicolumn{4}{c}{FaceScrub} & \multicolumn{4}{c}{CFW-60K} \\
        \cmidrule(lr){2-5}\cmidrule(lr){6-9}
        &\textbf{16-bit} &\textbf{24-bit} &\textbf{36-bit} &\textbf{48-bit} &\textbf{16-bit} &\textbf{24-bit} &\textbf{36-bit} &\textbf{48-bit} \\
        \midrule
        OPQN-A  &60.59  &87.62  &89.32  &89.89 &23.48  &74.97  &80.52  &81.72 \\
        OPQN-C  &61.64  &88.07  &83.90  &83.11 &54.55  &78.47  &73.84  &74.23 \\
        OPQN-S &20.16  &28.92  &60.72  &78.88  &16.86 &26.18  &62.15  &72.81 \\
        OPQN-W &80.66  &90.72  &92.40  &93.36  &73.38 &85.39  &87.89  &88.62 \\
        OPQN  &$\mathbf{90.32}$  &$\mathbf{91.54}$  &$\mathbf{92.70}$  &$\mathbf{93.85}$ &$\mathbf{85.47}$  &$\mathbf{86.37}$  &$\mathbf{88.26}$  &$\mathbf{89.40}$ \\
        \bottomrule
    \end{tabular}}
    \label{table:map_variant}
\end{table}
}

For a fair comparison, the same codeword configuration as in the original OPQN is applied to the four variants. 
We report the MAP results on FaceScrub and CFW-60K datasets in Table~\ref{table:map_variant}. One can see that all these methods exhibit different degrees of deterioration in performance as measured by MAP compared to OPQN. The comparison of OPQN-A and OPQN shows that involving original feature information in training is beneficial to obtain more discriminative representations. The improvement in performance from OPQN-C to OPQN verifies the advantage of discriminability maximization in each subspace for better precision in quantization. OPQN-S performs the worst among the four variants, indicating the necessity of $\ell_2$ normalization and the angular margin for the removal of radial variations and learning of separable representations. Finally, from the comparison of OPQN-W and OPQN, the entropy-based regularization boosts the performance under tiny code lengths, e.g., 16-bit, while for longer bits, our OPQN slightly performs better than OPQN-W. 
It indicates that disturbances from quantization errors may be more severe under short bit lengths. With longer bits, the adverse impact is lessened due to more possible combinations of codewords for quantization. 

\subsection{Codebook Configuration}
We further explore the influence of different codebook configurations on code performance. Basically, a $l$-bit binary code can be generated in the form of $l=M \times O$, where $M$ is the number of codebooks, $O=\log_2 K$, and $K$ is the number of codewords per codebook. Thus, binary codes with the same length can be obtained from different combinations of $M$ and $O$. For simplicity, we consider two relative configurations with $K$ ranging between $2^6$ and $2^9$: the first configuration adopts bigger $M$ and smaller $O$ for encoding while the other uses smaller $M$ and bigger $O$. Two sets of experiments on 24-bit, 36-bit, and 48-bit codes are conducted w.r.t two configurations. The combinations of the first configuration are $4 \times 6$, $6 \times 6$, and $8 \times 6$ for three code lengths, while the other configuration uses $3 \times 8$, $4 \times 9$, $6 \times 8$ as code lengths.

{\linespread{1.2}
\begin{table}[htbp]
    \centering
    \small
    \caption{Comparison of MAP ($\%$) results by different codebook configurations}
    \setlength{\tabcolsep}{1.2mm}{
    \begin{tabular}{ccccccc}
        \toprule
        \multirow{2}{*}{Dataset} & \multicolumn{2}{c}{24-bit} &  \multicolumn{2}{c}{36-bit} & \multicolumn{2}{c}{48-bit} \\ 
        \cmidrule(lr){2-3}\cmidrule(lr){4-5} \cmidrule(lr){6-7}
        & $\mathbf{4 \times 6}$ &$\mathbf{3 \times 8}$ &$\mathbf{6\times 6}$ &$\mathbf{4 \times 9}$ &$\mathbf{8 \times 6}$ &$\mathbf{6 \times 8}$ \\
        \midrule
        FaceScrub  &91.54  &$\mathbf{92.98}$  &92.70  &$\mathbf{93.67}$ &$\mathbf{93.85}$  &93.29  \\
        VGGFace2 &77.22  &$\mathbf{89.86}$  &90.26  &$\mathbf{95.08}$ &93.86  &$\mathbf{95.04}$  \\
        \bottomrule
    \end{tabular}}
    \label{table:map_config}
\end{table}
}

The standard retrieval results on FaceScrub and VGGFace2 under different codebook configurations are presented in Table~\ref{table:map_config}. We can see performance improvement in almost all the cases when using a larger $O$ for quantization. These improvements are more distinct for the VGGFace2 dataset. OPQN performs better with a larger value of $K$, especially for smaller bits. Using more codewords in a codebook implies richer prototypes in the subspace for quantization, making it easier for sub-vectors to find the nearest prototype with less quantization error. By using $K=256$, rather than $K=64$, under 24-bit codes on the VGGFace2 dataset, OPQN can achieve nearly a 90$\%$ MAP score and the performance of 36-bit codes with a $4 \times 9$ configuration is better than that with 48-bit codes. The benefit of a larger $K$ may become less with longer bit codes as the increase in the number of subspaces might reduce the quantization error.

\subsection{Parameter Sensitivity}
The effect of parameters on model performance is examined. From the ablation study shown in Table~\ref{table:map_variant}, entropy-based regularization mainly exhibits its advantage under shorter codes. Thus, we present MAP results w.r.t. different values of the balance weight, $\lambda$ under 16-bit codes in Table~\ref{table:map_lamda}. One can see that the performance increases steadily as the regularization with $\lambda$ varies from 0.01 to 0.1 but decreases sharply for values bigger than 0.1.

{\linespread{1.2}
\begin{table}[htbp]
    \centering
    \small
    \caption{16-bit MAP ($\%$) results w.r.t. different values of $\lambda$}
     \setlength{\tabcolsep}{1.5mm}{
    \begin{tabular}{cccccc}
        \toprule
       $\lambda$\  &0 &0.01 &0.05 &0.1 &0.3  \\
        \midrule
        FaceScrub  &80.66 &82.47  &85.43  &90.32  &1.59 \\
        CFW-60K   &73.38 &76.28  &78.01  &85.47  &10.29 \\
        \bottomrule
    \end{tabular}}
    \label{table:map_lamda}
\end{table}
}

\begin{figure}[htbp]
	\centering
	\includegraphics[width = 0.65\textwidth]{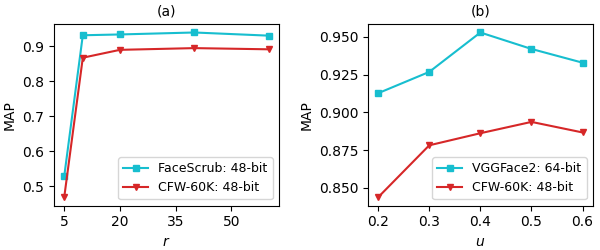}
	\caption{MAP results w.r.t. different values of (a) the scaling factor $r$, (b) the margin $u$, both under two experimental settings, respectively.}
	\label{param_sm}
\end{figure}
We then study the impacts of the scaling factor $r$ and the angular margin $u$, in Eqs.~(\ref{7}) and (\ref{8}). The plots of MAP performances w.r.t. different values of $r$ and $u$ are illustrated in Fig.~\ref{param_sm}(a) and (b), respectively. From Fig.~\ref{param_sm}(a), performance measured by MAP improves substantially from $r=5$ to $r=10$, reaching good values, and then is nearly stable. In Fig.~\ref{param_sm}(b), as $u$ increases from 0.2 to 0.6, the performance of the model rises and then decreases. The turning point appears earlier on the VGGFace2 dataset with the best values of $u$ being around 0.4 in VGGFace2 and 0.5 in the CFW-60K dataset. Generally, the model performs robustly under variations of $u$. 

\subsection{Advantage of Orthogonality}
One key feature of the proposed OPQN is that we use orthonormal codewords instead of other types of predefined codewords. We design this orthogonality accounting for three main reasons. Firstly, it is more feasible to produce orthonormal vectors than vectors with other specified angles in between. By using simple but effective DCT transform or SVD, one can easily obtain the desired codewords. Secondly, regarding improving codeword informativeness and reducing codeword redundancy, we expect codewords to scatter uniformly and retain a distance from each other. The orthogonality provides a decent separable property to predefined codewords. Lastly, orthogonality leads to simplifying the asymmetric distance comparison according to Eq.~\ref{20}. Consequently, it requires less computation cost than any other type of codewords. 

\begin{figure}[htbp]
	\centering
	\includegraphics[width = \textwidth]{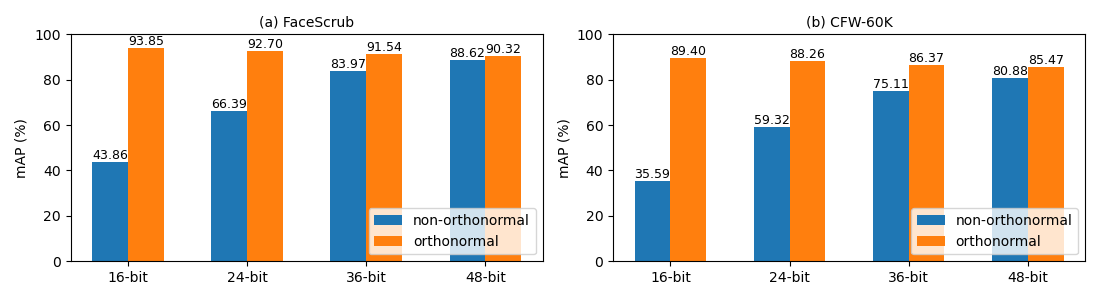}
	\caption{Comparison of using predefined codewords with and w/o orthogonality under four code lengths on two datasets.}
	\label{exp_ortho}
\end{figure}

One may wonder how is the performance when using predefined codewords without orthogonality. To this end, we add some random noise of $\mathcal{N}(0, 1e-4)$ to the generated orthonormal codewords as non-orthonormal codewords. Experiment results comparing the seen identity retrieval performance of these two kinds of codewords on FaceScrub and CFW-60K datasets are shown in Fig.~\ref{exp_ortho}. We can see that the permanence drops dramatically when breaking the orthogonality, even with a small variance of 1e-4. And the degradation is more significant with the decrease in the code length. The results validate the superiority of orthogonality for predefined codewords quantitatively, indicating the necessity of orthogonality under the proposed deep quantization framework. 

\section{Conclusion}\label{sec:6}
This paper develops a deep product quantization-based method, OPQN, particularly for face image retrieval. Unlike previous deep quantization works, a novel framework is proposed using predefined orthonormal codewords for quantization. A tailored loss function is designed as the learning metric to maximize discriminability in both the soft quantization and original features in all subspaces. Comprehensive and extensive experiments are conducted under both seen and unseen identity retrieval settings. OPQN outperforms a series of compared deep hashing/quantization methods under both settings. Its superior generalization performance induced by the proposed orthonormal codewords verifies the importance of codewords distribution to quantization quality.

Future work could include investigating how the visual information is scattered into subspaces in the PQ-based method. The sub-vector of bottleneck features in different subspaces may encode specific facial regions or attributes. Thus, it would be useful to design a learning metric that is adaptive to different parts of visual discriminability in the quantization subspace. 

\section*{Acknowledgement}
This work is supported by Hong Kong Innovation and Technology Commission (InnoHK Project CIMDA), Hong Kong Research Grants Council (Project 11204821), and City University of Hong Kong (Project 9610034).
\footnotesize
{\linespread{1.3}
{\bibliography{reference}}
}
\end{document}